\documentclass[eat,twocolumn]{jmlr}
   
\usepackage{longtable}

\usepackage{booktabs}
\usepackage[load-configurations=version-1]{siunitx} 
\usepackage{dsfont}

\theorembodyfont{\upshape}
\theoremheaderfont{\scshape}
\theorempostheader{:}
\theoremsep{\newline}

\jmlrvolume{1}
\firstpageno{1}

\jmlryear{2022}
\jmlrworkshop{Machine Learning for Health (ML4H) 2022}

\title[Pipeline-Invariant Representation Learning for Neuroimaging]{Pipeline-Invariant Representation Learning for Neuroimaging}

  \author{\Name{Xinhui Li}$^1$\thanks{Corresponding authors} \Email{xinhuili@gatech.edu}\\
  \Name{Alex Fedorov}$^1$\footnotemark[1] \Email{afedorov@gatech.edu}\\
  \Name{Mrinal Mathur}$^1$ \Email{mmathur4@student.gsu.edu}\\
  \Name{Anees Abrol}$^1$ \Email{aabrol@gsu.edu}\\
  \Name{Gregory Kiar}$^2$ \Email{gregory.kiar@childmind.org}\\
  \Name{Sergey Plis}$^1$ \Email{splis@gsu.edu}\\
  \Name{Vince Calhoun}$^1$ \Email{vcalhoun@gsu.edu}\\
  \addr $^1$The Georgia State University/Georgia Institute of Technology/Emory University Center for Translational Research in Neuroimaging and Data Science (TReNDS), Atlanta, GA, USA\\
  \addr $^2$Child Mind Institute, New York, NY, USA}

\begin{document}

\maketitle

\begin{abstract}
Deep learning has been widely applied in neuroimaging, including predicting brain-phenotype relationships from magnetic resonance imaging (MRI) volumes.  MRI data usually requires extensive preprocessing prior to modeling, but variation introduced by different MRI preprocessing pipelines may lead to different scientific findings, even when using the identical data. Motivated by the data-centric perspective, we first evaluate how preprocessing pipeline selection can impact the downstream performance of a supervised learning model. We next propose two pipeline-invariant representation learning methodologies, \emph{MPSL} and \emph{PXL}, to improve robustness in classification performance and to capture similar neural network representations. Using $2000$ human subjects from the UK Biobank dataset, we demonstrate that proposed models present unique and shared advantages, in particular that MPSL can be used to improve out-of-sample generalization to new pipelines, while PXL can be used to improve within-sample prediction performance. Both MPSL and PXL can learn more similar between-pipeline representations. These results suggest that our proposed models can be applied to mitigate pipeline-related biases, and to improve prediction robustness in brain-phenotype modeling.
\end{abstract}

\section{Introduction}
\label{sec:intro}

Deep learning has been widely applied to establish novel brain-phenotype relationships and to advance our understanding of brain disorders, in part because of its effectiveness in learning nonlinear relationships from neuroimaging data (e.g., magnetic resonance imaging;
MRI)~\citep{plis2014deep,abrol2021}. MRI data usually requires extensive preprocessing to mitigate data collection artifacts and transform the data to standard spaces for performing statistical analyses and interpretation of results. In the past decade, a growing array of MRI preprocessing pipelines have been developed, but there remains no consensus standard for preprocessing methods. Though these pipelines share basic preprocessing components, the specific implementation at each step can be different. Recent studies have shown that pipeline-related variation may result in significantly different preprocessed results and may lead to conflicting scientific conclusions, even when using identical raw data~\citep{botvinik2020,li2021}. When used in the development of deep learning models, these pipeline-specific biases may be amplified if models learn shortcut strategies based on unique non-biological features \citep{torralba2011,geirhos2020shortcut}. However, there is little work in the literature evaluating how preprocessing pipelines will affect downstream deep learning task performance.

Recently, the machine learning community has emphasized the importance of shifting from \textit{model-centric} to \textit{data-centric} approaches given that data quality plays an essential role in deep learning applications \citep{ng2021}. Motivated by this data-centric perspective, we first evaluate how preprocessed data from different pipelines affect the downstream performance of a supervised learning model. To this end, a uni-pipeline supervised learning (UPSL) model is trained, using a dataset preprocessed by each of three pipelines, respectively. 
We train these models on a challenging combined age and gender classification task from a previous study \citep{abrol2021} to investigate how the model performance is sensitive to pipeline-related variation. We then compare models trained across pipelines through 1) within-sample test accuracy, 2) mutual agreement between pipelines, 3) out-of-sample test accuracy from transfer learning and 4) representational similarity of convolutional layers measured by minibatch centered kernel alignment (CKA) \citep{nguyen2020wide}. Our results highlight significant pipeline-related variation and poor generalizability in UPSL.

Next, we propose two approaches to mitigate pipeline-related variation and learn pipeline-invariant representations. First, we suggest a multi-pipeline supervised learning (MPSL) model trained on dataset pairs to take features from both datasets into account. Second, we introduce a pipeline-based contrastive learning (PXL) model which integrates both supervised and contrastive learning paradigms. These approaches were evaluated similarly to the UPSL models, and our findings demonstrate that both techniques have unique strengths. Specifically, MPSL can improve out-of-sample generalization to new pipelines, while PXL can achieve competitive and consistent performance within a pipeline set. Notably, both MPSL and PXL can improve latent representational similarity. 

The key contributions of this study include: 1) evaluation of the impact of neuroimaging preprocessing pipelines on deep learning tasks; 2) proposal of novel methodologies to evaluate learning performance including mutual agreement between pipelines, within-sample and out-of-sample test accuracy, and between-pipeline CKA; 3) development of two pipeline-invariant representation learning methodologies, \emph{MPSL} and \emph{PXL}, to capture pipeline-invariant representations in the latent space and mitigate pipeline-related variation in prediction tasks, including when applied to out-of-sample pipelines.

\section{Methods}

\subsection{Data Preprocessing}
The T1-weighted structural MRI (sMRI) images of $2000$ subjects from UK Biobank dataset \citep{miller2016multimodal, abrol2021} were used in this study. Subjects were grouped into $5$ age groups ($45-52$, $53-59$, $60-66$, $67-73$, and $74-80$ years old) and $2$ sex groups (male and female), resulting in $10$ labels in total. The $2000$ subjects were selected to balance the age and sex categories in the dataset. $1800$ subjects with balanced labels were randomly selected and then evenly split into $9$ folds for hyperparameter optimization and cross-validation. The remaining $200$ subjects with balanced labels were used as a hold-out test set. We report the inference performance on the hold-out test set from models trained on $9$ folds.

The same sMRI dataset was preprocessed by each of three commonly-used MRI preprocessing pipelines independently: the default pipeline in the Configurable Pipeline for the Analysis of Connectomes (C-PAC:Default) \citep{craddock2013}, the fMRIPrep-options pipeline in C-PAC (C-PAC:fMRIPrep) \citep{esteban2019fmriprep}, and the UK Biobank FSL pipeline followed by SPM (UKB FSL-SPM) \citep{alfaro2018image, jenkinson2012fsl, friston1994statistical}. The detailed preprocessing workflow is described in Appendix \ref{appendix_workflow}.
The preprocessed gray matter volume image, a known biomarker of aging and gender effects \citep{silva2021direct}, was used as the input. 

   \begin{figure}[tpb]
      \centering
      \includegraphics[width=3in]{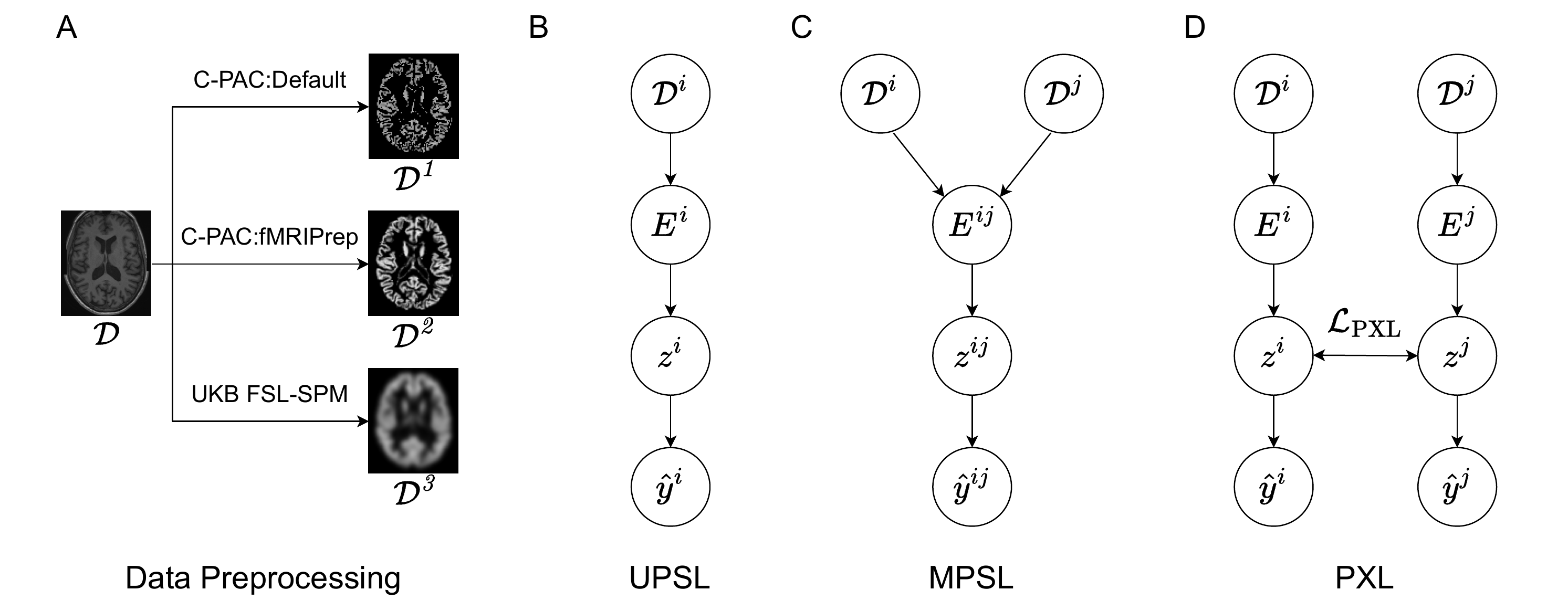}
      \caption{Experiment overview.}
      \label{overview}
   \end{figure}

\subsection{Model Architectures}


\textbf{Uni-Pipeline Supervised Learning}~~
To evaluate how data preprocessing affects the prediction result, we trained a supervised learning model in a combined age and gender prediction task for each preprocessed dataset separately, denoted as uni-pipeline supervised learning (UPSL).
The UPSL model includes one encoder $E^i$, taking each of three datasets $\mathcal{D}^i$ as the input, learning representations $z^i$ and predicting labels $\hat{y}^{i}$. The encoder network was developed based on AlexNet \citep{krizhevsky2012} because it is widely-used in the neuroimaging literature \citep{lin2021utilizing, zhang2020three,fedorov2019prediction} and previous work \citep{abrol2021} provides a performance benchmark. The AlexNet architecture is described in Appendix \ref{appendix_alexnet}. 
Each model was then trained for $200$ epochs. We repeated the experiment across $9$ folds of training and validation data.

\textbf{Multi-Pipeline Supervised Learning}~~
Our first proposed architecture, multi-pipeline supervised learning (MPSL), includes one encoder $E^{ij}$ taking two datasets $\mathcal{D}^{i}$ and $\mathcal{D}^{j}$ to learn representations $z^{ij}$ and predict labels $\hat{y}^{ij}$.
The idea of MPSL is to treat pipelines as unique data augmentation transformations.
Such strategy doubles the size of training data, but the training process and the model implementation are identical to UPSL. 

\textbf{Pipeline-based Contrastive Learning}~~
Our second proposed approach, pipeline-based contrastive learning (PXL), consists of two encoders ($E^i$, $E^j$) using the dataset preprocessed by two pipelines ($\mathcal{D}^i$, $\mathcal{D}^j$) as the inputs separately, and each producing their own sets of output labels ($\hat{y}^i$ and $\hat{y}^j$). 
The novel contribution in PXL is to add a contrastive loss term to the supervised loss function to bring the representations from different pipelines closer to each other in the latent space for the same subject, while pushing away the representations for different subjects. 
The details of the PXL contrastive objective $\mathcal{L}_\mathrm{PXL}$ are explained in Appendix \ref{appendix_loss} and \ref{appendix_lambda}. 

The hyperparameter optimization for each of three models is described in Appendix \ref{appendix_hypopt}. 

\subsection{Evaluation Metrics}
We use three metrics to measure pipeline-invariant learning performance: within-sample test accuracy, mutual agreement across pipelines, and out-of-sample test accuracy. The within-sample test accuracy was obtained by applying the trained model on the hold-out test set preprocessed by the same pipeline. The mutual agreement across pipelines was calculated as the percentage of overlap between the predicted labels from a pipeline pair $\hat{y}^i$, $\hat{y}^j$ and the ground truth labels $y$ (i.e. $\hat{y}^i=\hat{y}^j=y$). 
To evaluate out-of-sample generalizability, we trained a logistic regression model from \emph{scikit-learn}~\citep{scikit-learn} using the training set from a different pipeline. 
The pretrained encoder with fixed parameters served as a feature extractor and the learned representations were preserved during transfer learning. Minibatch CKA was used to measure representational similarity between pipeline pairs (see Appendix \ref{appendix_minicka}).

\section{Results}

\subsection{PXL achieves competitive within-sample performance while MPSL demonstrates robust out-of-sample generalization.}

\begin{figure}[tpb]
  \centering
  \includegraphics[width=3in]{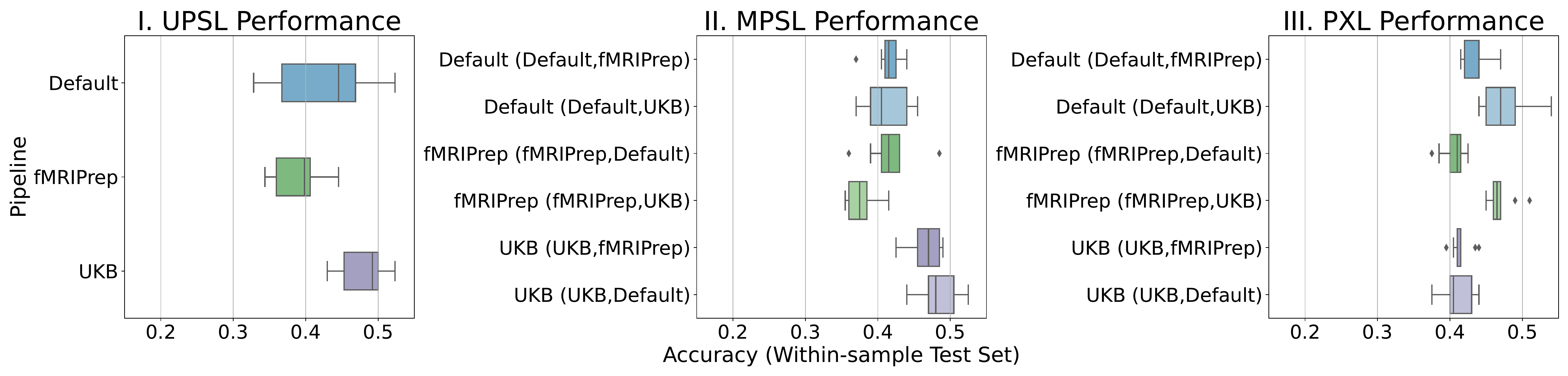}
  \caption{Within-sample test performance.}
  \label{infer}
\end{figure}

The UPSL average test accuracy ranges from $0.390$ to $0.482$, with a difference of $0.092$ (Figure \ref{infer}\footnote{Default, fMRIPrep and UKB stand for C-PAC:Default, C-PAC:fMRIPrep and UKB FSL-SPM, respectively. The training set pair is indicated in brackets.} and Table \ref{table_infer}\footnote{MPSL and PXL training set pair order matches the vertical $y$-axis label order in Figure \ref{infer}.}). 
The statistical analysis reveals that the test result from UKB FSL-SPM is significantly different from the results from C-PAC:Default and C-PAC:fMRIPrep ($p < 0.05/3$, Wilcoxon signed-rank test, Bonferroni correction). The performance difference between the C-PAC:Default and C-PAC:fMRIPrep pipelines is not significant. 
We further replicated the UPSL experiment using the DCGAN encoder \citep{radford2015unsupervised}, an effective unsupervised representation learning encoder, and observed a similar performance difference of $0.069$ across pipelines (see Appendix \ref{upsl_dcgan}). The UPSL result indicates that preprocessed data from different pipelines will significantly affect the downstream prediction performance. 

\begin{table}[tpb]
\tiny
  \caption{Within-sample test performance.}
  \label{table_infer}
  \begin{tabular}{cccc}
  \toprule
  Model & Default & fMRIPrep & UKB\\
  \midrule
  UPSL & $0.420 \pm 0.064$ & $0.390 \pm 0.032$ & $\mathbf{0.482 \pm 0.030}$ \\
  MPSL & $0.416 \pm 0.035$ & $0.424 \pm 0.018$ & $0.460 \pm 0.032$ \\
      & $0.398 \pm 0.015$ & $0.347 \pm 0.033$ & $0.448 \pm 0.016$ \\
  PXL & $0.437 \pm 0.019$ & $0.406 \pm 0.016$ & $0.415 \pm 0.013$ \\
      & $\mathbf{0.474 \pm 0.030}$ & $\mathbf{0.469 \pm 0.018}$ & $0.407 \pm 0.022$ \\
  \bottomrule
  \end{tabular}
\end{table}

In MPSL, the overall performance is more consistent, except that the C-PAC:fMRIPrep test set result from the C-PAC:fMRIPrep and UKB FSL-SPM training set pair is lower than the others.
The MPSL result shows that pipeline-related variation can be mitigated by incorporating multiple preprocessed datasets during training. 
In PXL, the performance difference ($0.068$) is the smallest among three models, with smaller cross-fold variance observed. Interestingly, the inference performance on the C-PAC:Default and C-PAC:fMRIPrep test sets becomes significantly better when the encoder utilizes the UKB FSL-SPM dataset during training, though the inference performance on the UKB FSL-SPM test set slightly drops compared to UPSL and MPSL.
Both MPSL and PXL demonstrate higher mutual agreement than UPSL (Table \ref{table_agreement}). Specifically, PXL shows the best performance for the C-PAC:fMRIPrep and UKB FSL-SPM pair while MPSL is the best for the other pairs.

\begin{table}[tpb]
\tiny
  \caption{Mutual agreement.}
  \label{table_agreement}
  \begin{tabular}{cccc}
  \toprule
  Model & \begin{tabular}{@{}c@{}}Default \\fMRIPrep\end{tabular} & \begin{tabular}{@{}c@{}}Default \\UKB\end{tabular} & \begin{tabular}{@{}c@{}}fMRIPrep \\UKB\end{tabular} \\
  \midrule
  UPSL & $0.206 \pm 0.030$ & $0.228 \pm 0.043$ & $0.228 \pm 0.014$ \\
  MPSL & $\mathbf{0.233 \pm 0.028}$ & $\mathbf{0.259 \pm 0.022}$ & $0.228 \pm 0.017$ \\
  PXL & $0.217 \pm 0.016$ & $0.226 \pm 0.028$ & $\mathbf{0.246 \pm 0.015}$ \\
  \bottomrule
  \end{tabular}
\end{table}

\begin{table}[tpb]
\tiny
  \caption{Out-of-sample transfer learning test performance.}
  \label{table_generalizability}
  \begin{tabular}{cccc}
  \toprule
  Model & Default & fMRIPrep & UKB \\
  \midrule
UPSL & - & $0.265 \pm 0.030$ & $0.256 \pm 0.024$ \\
& $\mathbf{0.297 \pm 0.027}$ & - & $0.266 \pm 0.019$ \\
& $0.222 \pm 0.023$ & $0.207 \pm 0.024$ & - \\
MPSL & $0.278 \pm 0.032$ & $0.285 \pm 0.014$ & $0.282 \pm 0.027$ \\
PXL & $0.266 \pm 0.043$ & $\mathbf{0.290 \pm 0.031}$ & $\mathbf{0.309 \pm 0.019}$ \\
& $0.219 \pm 0.031$ & $0.216 \pm 0.029$ & $0.288 \pm 0.030$ \\
  \bottomrule
  \end{tabular}
\end{table}

The out-of-sample transfer learning performance, which measures model generalizability, is presented in Table \ref{table_generalizability}\footnote{In UPSL, each row corresponds to Default, fMRIPrep and UKB training set, respectively. In MPSL and PXL, each column corresponds to fMRIPrep - UKB, Default - UKB, and Default - fMRIPrep training set pair, respectively.}. We observe the best generalization performance from MPSL models (average across all models of $0.282$), as well as fairly good performance from the PXL models (average across all models of $0.265$). UPSL performance is not optimal when transferred to an unseen test set (average across all models of $0.252$), suggesting the learned features are not guaranteed to generalize across pipelines.

\subsection{MPSL and PXL capture more similar between-pipeline representations to UPSL.}
\label{results_cka}

\begin{figure}[tpb]
      \centering
      \includegraphics[width=3in]{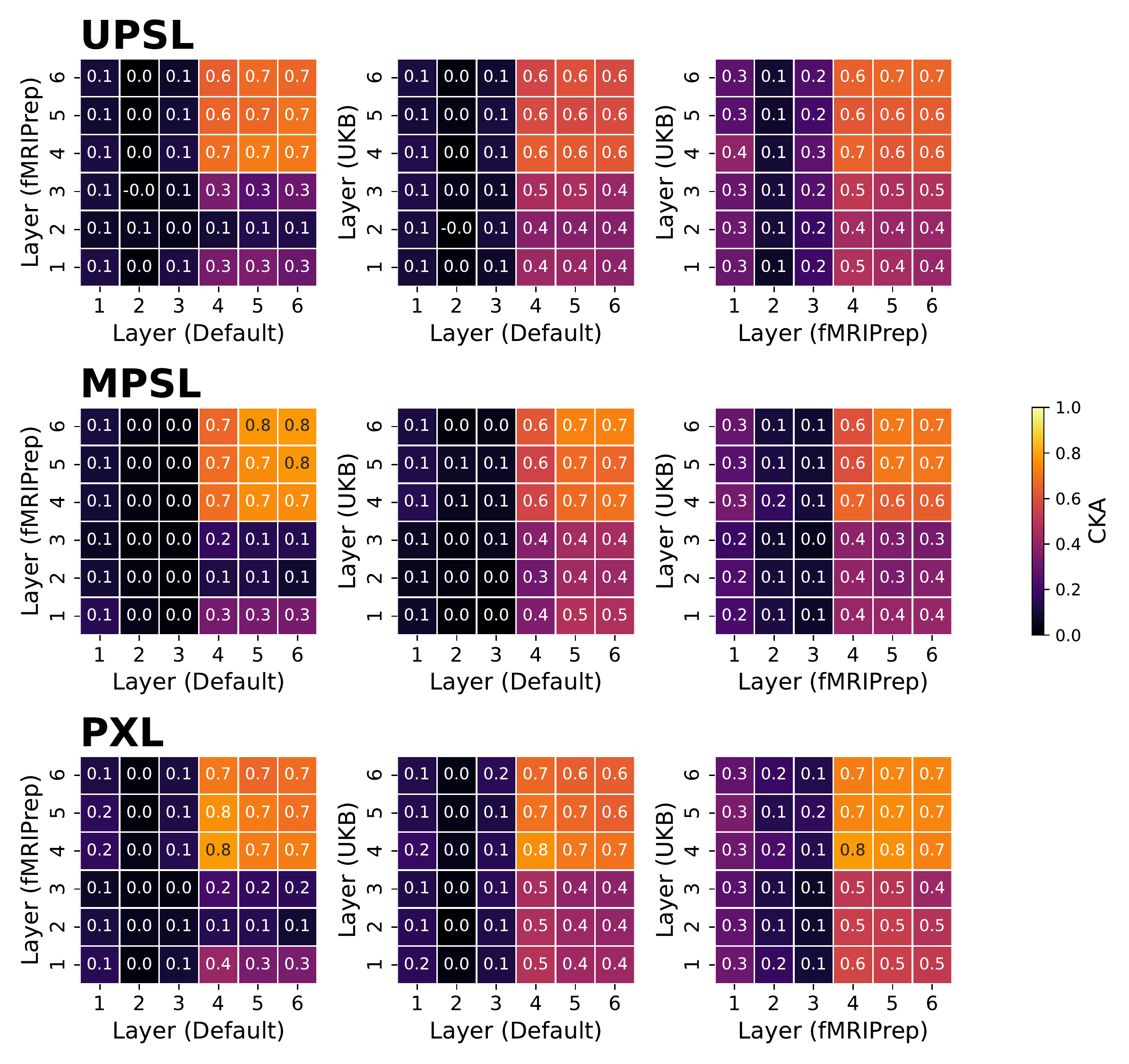}
      \caption{Between-pipeline CKA.}
      \label{cka}
  \end{figure}
\begin{figure}[tpb]
      \centering
      \includegraphics[width=3in]{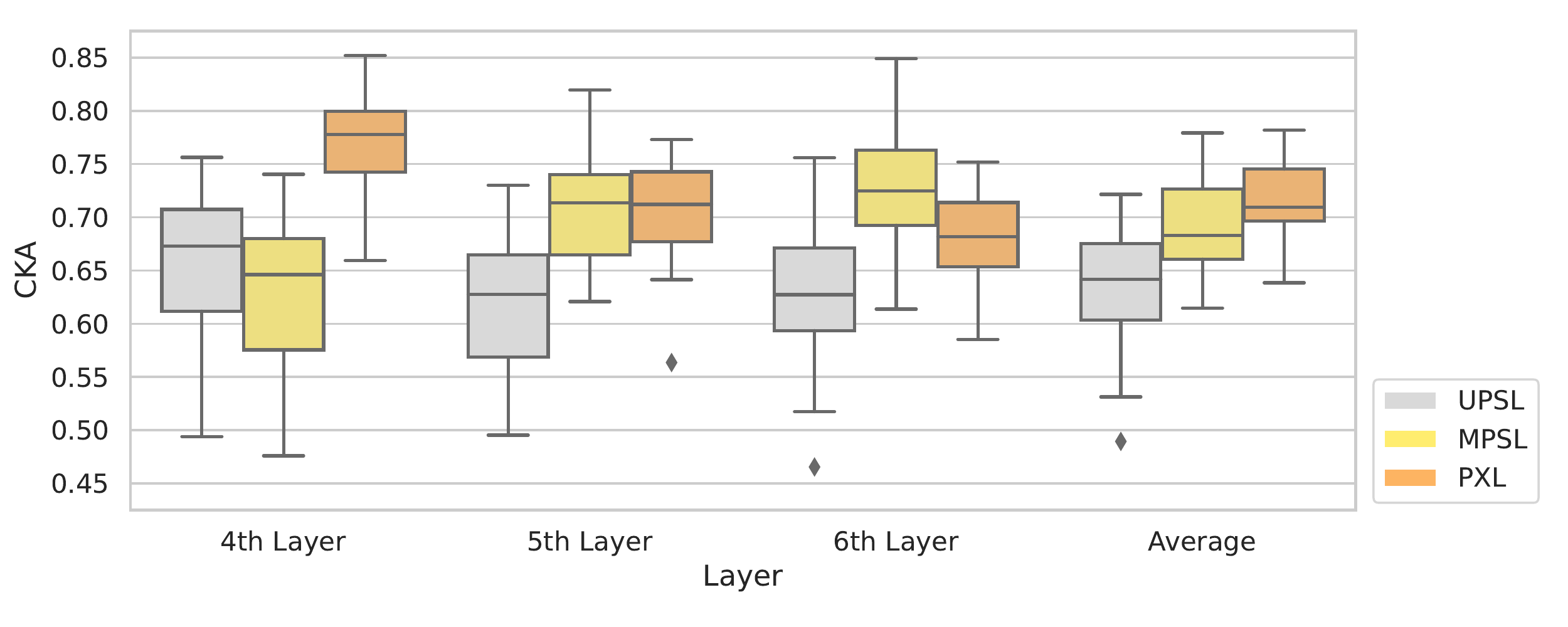}
      \caption{Between-pipeline CKA in the last three layers.} 
      \label{cka_boxplot}
   \end{figure}

As shown in Figure \ref{cka}, between-pipeline representations from the last three layers are more similar while those from the first three layers are less similar. Both MPSL and PXL have higher CKA values than UPSL, suggesting the learned representations from MPSL and PXL are more similar than UPSL. Among three models, PXL has the highest average between-pipeline CKA value $0.718$ across the last three layers (see Appendix \ref{appendix_cka} Table \ref{cka_table}). Figure \ref{cka_boxplot} also demonstrates that highly similar between-pipeline representations learned by MPSL and PXL in the last three layers. Further, we performed experiments on natural image datasets and showed that PXL can improve representational similarity compared to UPSL (see Appendix \ref{appendix_mnist}).


\section{Discussion}

The contributions of the present work are two-fold. First, we evaluated the impact of neuroimaging preprocessing pipelines and preprocessed data quality in deep learning tasks, and demonstrated current limitations of UPSL. The UPSL result demonstrates that the same dataset preprocessed by different pipelines can result in significantly different prediction performance. As shown in Figure \ref{infer} and Table \ref{table_infer}, the difference of UPSL within-sample test accuracy from different pipelines can be as high as $9.2\%$ (chance accuracy: $10\%$) when using datasets preprocessed by different pipelines. 
Additionally, UPSL models cannot generalize well to unseen pipelines. 
The result emphasizes the importance of clear scientific communication surrounding decisions in neuroimaging preprocessing, and making pipelines publicly available to allow for evaluation, comparison, and reproduction in the context of downstream learning tasks. 

Next, we proposed two approaches, MPSL and PXL, to mitigate pipeline-related variation. While the MPSL approach is a naive extension of UPSL, we noted that the MPSL approach led to consistent within-sample performance while also improving out-of-sample generalizability. We hypothesize this is because the MPSL model was less prone to learning pipeline-specific features, and pipeline differences forced the optimization to identify features that discriminated participants across more biologically-meaningful features. 
Our novel approach PXL adopts a contrastive loss function leading to the improved within-sample performance. 
Specifically, PXL achieved the highest within-sample test accuracy on the C-PAC:Default and C-PAC:fMRIPrep datasets, though the UKB FSL-SPM performance dropped relative to UPSL and MPSL (Table \ref{table_infer}). 
One possibility is that the contrastive objective in PXL learns the shared information from both views and may tend to capture high-frequency texturized features but not low-frequency information from UKB FSL-SPM, the only pipeline which applies spatial smoothing. 
UKB FSL-SPM smoothed out texturized information that includes individual variability and thus achieved the best within-sample test performance in UPSL.
Notably, both MPSL and PXL capture more similar representations in the last three layers (Figures \ref{cka}, \ref{cka_boxplot}), supporting their potentials to achieve pipeline-invariant learning. 

Future work will evaluate these approaches on other neuroimaging modalities such as functional MRI that incorporates temporal dynamics as well as on a wider range of tasks including brain disorder prediction. 
Furthermore, MPSL and PXL can be applied to mitigate site effects or data acquisition effects. The experiments on natural image datasets (Appendix \ref{appendix_mnist}) support the feasibility of pairing samples by labels in the contrastive learning paradigm.

In summary, we demonstrated the pipeline-related variation can make a significant difference in the prediction result of a downstream task. We then proposed two pipeline-invariant representation learning approaches, MPSL and PXL, to mitigate the biases introduced by data preprocessing. Our results demonstrated that MPSL and PXL can achieve robust and consistent within-sample and out-of-sample inference performance and improve representational similarity in the latent space. The proposed models can be applied to mitigate pipeline-related variation and improve prediction robustness in brain-phenotype modeling. 

\acks{This work was supported by NSF grant \#2112455, NIH grant R01MH118695 and NIH/NIBIB grant T32EB025816.}

\bibliography{pmlr-sample}

\appendix

\setcounter{figure}{0}
\counterwithin{figure}{section}
\setcounter{table}{0}
\counterwithin{table}{section}

\section{Preprocessing workflows}\label{appendix_workflow}

The detailed preprocessing workflow of each pipeline is as follows: 1) The \emph{C-PAC:Default} structural preprocessing workflow performs brain extraction via AFNI 3dSkullStrip \citep{cox1996afni}, tissue segmentation via FSL FAST \citep{zhang2001segmentation}, and spatial normalization via ANTs SyN non-linear alignment \citep{avants2008symmetric}. 2) The \emph{C-PAC:fMRIPrep} structural pipeline applies ANTs N4 bias field correction \citep{tustison2010n4itk} on the raw images, followed by ANTs brain extraction, a custom thresholding and erosion algorithm to generate tissue segmentation masks \citep{esteban2019fmriprep}, and ANTs SyN alignment to transform the data to the standard space. ANTs registration is performed using skull-stripped images, unlike the C-PAC:Default pipeline which uses whole-head images. 3) The \emph{UKB FSL-SPM} pipeline runs a gradient distortion correction and calculates linear and non-linear transformations via FSL FLIRT \citep{jenkinson2001global, jenkinson2002improved} and FNIRT \citep{andersson2007a, andersson2007b}, respectively. Then it performs brain extraction via FSL BET \citep{smith2002fast} and segments the sMRI data into tissue probability maps. The gray matter images are then warped to standard space, modulated and smoothed using a Gaussian kernel with an FWHM $= 10$ mm using SPM12 \citep{friston1994statistical}.

All preprocessed gray matter volume images are in MNI (2006) space \citep{grabner2006symmetric}. The dimensions of the gray matter volume image are $91 \times 109 \times 91$, corresponding to a voxel size of $2 \times 2 \times 2$ mm$^3$.

\section{AlexNet architecture}\label{appendix_alexnet}
The AlexNet encoder includes $5$ convolutional layers and $1$ average pooling layer. The $5$ convolutional layers have $64$, $128$, $192$, $192$, $64$ output units, and $62^3$, $18^3$, $6^3$, $6^3$, $6^3$ output dimensions, respectively. The last convolutional layer with $64$ output units defines a $64$ dimensional representation.

\section{Contrastive loss in PXL}\label{appendix_loss}

We hypothesize that preprocessing pipelines may introduce unintended biases to the dataset and thus result in disagreement of learned representations and model performance. To mitigate pipeline-related biases, we aim to learn pipeline-invariant representations by maximizing agreement between differently preprocessed views of data. 

The goal of contrastive representation learning is to learn a latent space where representations of similar sample pairs are closer while representations of dissimilar ones are further apart. The PXL architecture is inspired by recent work on contrastive learning. Early applications of contrastive loss can be traced back to learning invariant mappings of certain input transformations \citep{chopra2005learning, hadsell2006dimensionality}. Several recent advances are rooted in Noise Contrastive Estimation (NCE) \citep{gutmann2010} including Memory Bank \citep{wu2018unsupervised}, Contrastive Predictive Coding \citep{van2018representation} and Deep InfoMax \citep{hjelm2018learning}. More recently, SimCLR \citep{chen2020} was proposed to train an encoder network and a projection head to maximize agreement between different views of the identical data via a contrastive loss. Moreover, self-supervision can improve model robustness when combining with a supervised loss \citep{hendrycks2019using}. PXL is developed by adding a contrastive loss to the supervised loss. The contrastive loss, defined by the NCE lower bound, is trained to bring representations from the same subject closer while pushing away those from different subjects in the latent space. Thus, PXL aims to learn pipeline-invariant latent representations.

The details of the contrastive objective in PXL are explained as below. Let $\mathcal{D}=\{(x^i, x^j; y) \sim (\mathcal{D}^i, \mathcal{D}^j)\}$ be a dataset of paired samples $(x^i, x^j; y)$, where $x^i$ is an input image from one dataset $\mathcal{D}^i$, $x^j$ is an input image from another dataset $\mathcal{D}^j$, and $y$ is a class label. Then we learn two independent encoders $E^i$ and $E^j$ parameterized by convolutional neural networks that map input images $x^i$ and $x^j$ to representations 
$z^i = E^i(x^i)$ and $z^j = E^j(x^j)$. 
To learn the parameters of the encoders, we optimize the PXL objective:
\begin{equation}
        \mathcal{L}_\mathrm{PXL} = \lambda \cdot  \ell_\mathrm{supervised} + (1 - \lambda) \cdot \ell_\mathrm{contrastive},
\end{equation}
where $\lambda$ is trade-off hyperparameter between the supervised and contrastive loss. Note that the approach can become self-supervision when $\lambda=0$ and it can also turn to a fully supervised model by setting $\lambda=1$, equivalent to MPSL with two encoders. We evaluate how the choice of $\lambda$ affects model performance and representational similarity in Appendix \ref{appendix_lambda}. The supervised loss $\ell_\mathrm{supervised}$ is defined as as sum of cross-entropy losses $\ell_\mathrm{CE}$ for pipeline $i$ and $j$:
\begin{equation}
        \ell_\mathrm{supervised} = \ell_\mathrm{CE}(g^i(z^i); y) + \ell_\mathrm{CE}(g^j(z^j); y),
\end{equation}
where $g$ is a linear projection head from representations to class labels. The contrastive loss term $\ell_\mathrm{contrastive}$ follows the Noise Contrastive Estimation (NCE) lower bound definition by~\citep{gutmann2010}.
For the $n$-th sample with a positive pair $(z^i_n, z^j_n)$, the contrastive objective from pipeline $i$ to pipeline $j$ is:
\begin{equation}
        \ell_{i \rightarrow j}(z_n^i, z_n^j) = -\log \frac{e^{f(h^i_n, h^j_n)}}{\sum_{m=1}^{N} \mathds{1}_{[m \neq n]} e^{f(h^i_n, h^j_m)}},
\end{equation}
where $N$ is the total number of training subjects, $f$ is a critic function and $h^i$ is a projection head for pipeline $i$, $h_n^i = h^i(z_n^i)$~\citep{chen2020}. For the choice of critic $f$, we use scaled dot product and proposed regularization techniques as $L_2$ penalty and soft hyperbolic tangent ($tanh$) clipping of the critic scores \citep{bachman2019learning}. The contrastive loss is calculated in both directions to ensure its symmetry.

The final contrastive objective is defined as:
\begin{equation}
        \ell_\mathrm{contrastive} = \ell_{i \rightarrow j} + \ell_{j \rightarrow i}.
\end{equation}

\section{Evaluation of the trade-off parameter in the PXL objective}\label{appendix_lambda}
We vary the trade-off parameter $\lambda$ between the supervised and contrastive loss in the PXL objective $\mathcal{L}_\mathrm{PXL}$ ($\lambda = 0, 0.25, 0.5, 0.75, 1$) and evaluate how $\lambda$ affects the model performance and the representational similarity.

\begin{figure}[tpb]
  \centering
  \includegraphics[width=3in]{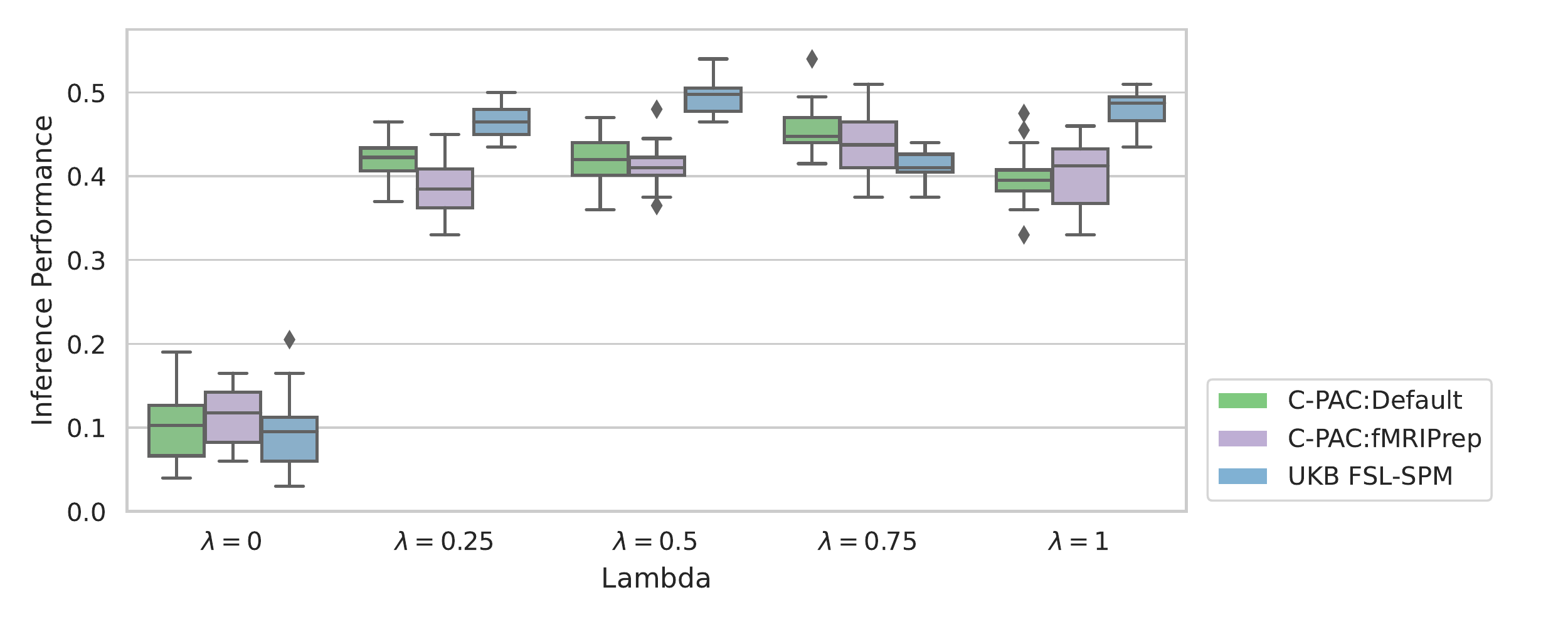}
  \caption{\textbf{Within-sample inference performance for different trade-off parameters $\lambda$ in the contrastive objective.} The box plot shows the within-sample test accuracy across $3$ models from $9$ cross-validation derived folds.}
  \label{appendix_pxl_infer}
\end{figure}

As shown in Figure \ref{appendix_pxl_infer} and Table \ref{appendix_table_generalizability}, we note that the inference performance is better when a larger $\lambda$ value is applied. Note that there is no classification head being trained when $\lambda=0$, which explains why the performance is relatively poor. The performance slightly improves when $\lambda$ increases from $0.25$ to $1$, implying that the supervised loss plays a more important role in improving the inference performance. 

\begin{figure}[tpb]
      \centering
      \includegraphics[width=3in]{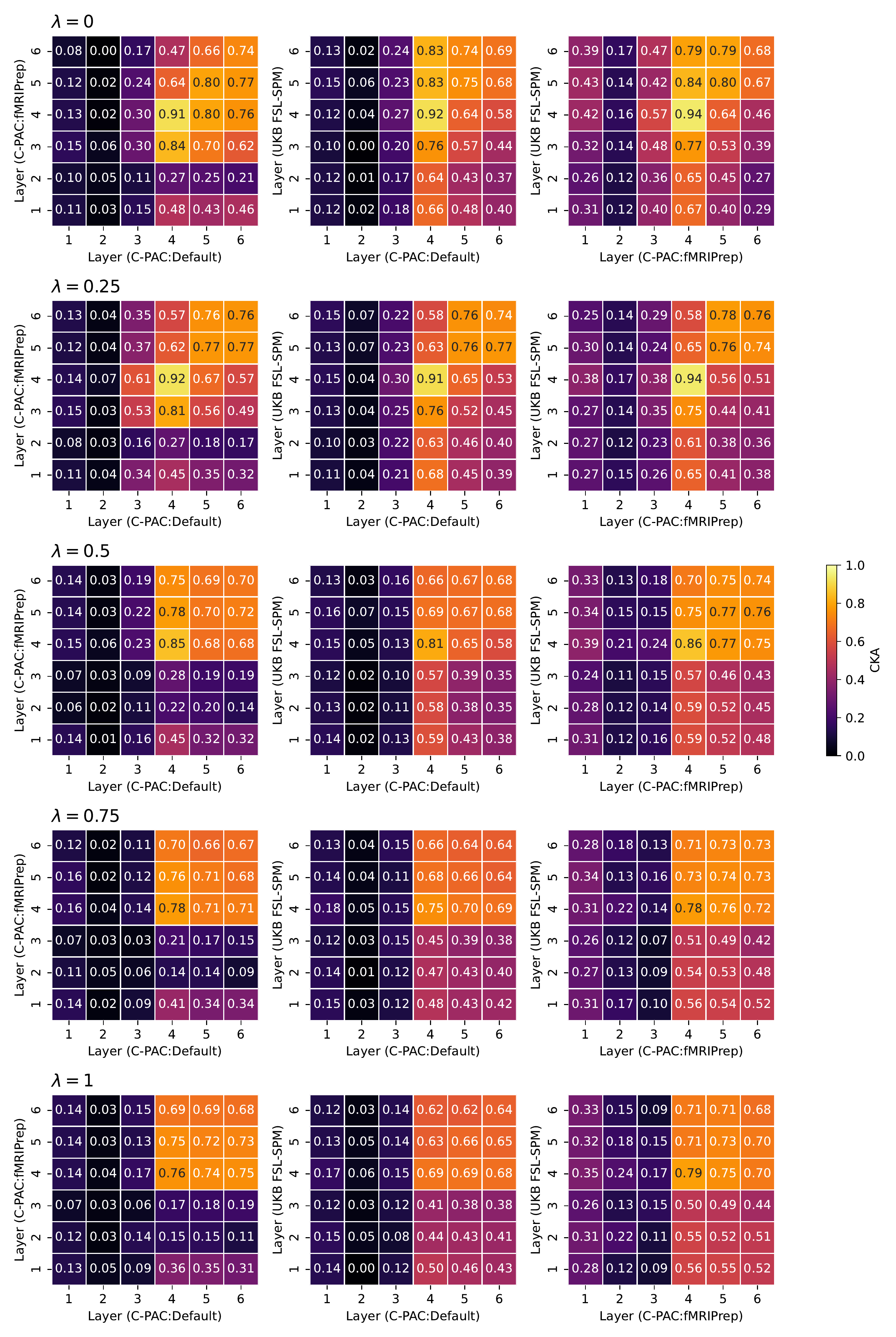}
      \caption{\textbf{PXL between-pipeline CKA across layers.} The average CKA across models from $9$ folds shows the last three layers share more similar representations in the hold-out test set. When $\lambda$ is smaller, the contrastive loss takes a larger percentage of the PXL objective, leading to more similar representations in the last three layers.}
      \label{appendix_pxl_cka}
  \end{figure}

According to Figure \ref{appendix_pxl_cka}, we observe that CKA values in the last three layers are negatively correlated with the $\lambda$ value, suggesting that the contrastive loss is the key to improve between-pipeline representational similarity.

\begin{table}[tpb]
\tiny
  \caption{\textbf{Out-of-sample transfer learning test performance for different trade-off parameters $\lambda$ in the contrastive objective.} Each column is evaluated at the training set pair C-PAC:fMRIPrep - UKB FSL-SPM, C-PAC:Default - UKB FSL-SPM, and C-PAC:Default - C-PAC:fMRIPrep, respectively.}
  \label{appendix_table_generalizability}
  \begin{tabular}{cccc}
  \toprule
  $\lambda$ & Default & fMRIPrep & UKB \\
  \midrule
  0 &
    \begin{tabular}{@{}c@{}}$0.178\pm0.087$\\
    $0.165\pm0.036$\end{tabular} & 
    \begin{tabular}{@{}c@{}}$0.196\pm0.071$ \\
    $0.186\pm0.063$\end{tabular} &
    \begin{tabular}{@{}c@{}}$0.188\pm0.066$ \\
    $0.172\pm0.081$\end{tabular} \\
    \midrule
    0.25 &
    \begin{tabular}{@{}c@{}}$0.237\pm0.126$\\
    $0.134\pm0.039$\end{tabular} & 
    \begin{tabular}{@{}c@{}}$0.260\pm0.070$ \\
    $0.202\pm0.062$\end{tabular} &
    \begin{tabular}{@{}c@{}}$0.261\pm0.065$ \\
    $0.207\pm0.097$\end{tabular} \\
    \midrule
    0.5 &
    \begin{tabular}{@{}c@{}}$0.175\pm0.072$\\
    $0.154\pm0.049$\end{tabular} & 
    \begin{tabular}{@{}c@{}}$0.212\pm0.103$ \\
    $0.179\pm0.054$\end{tabular} &
    \begin{tabular}{@{}c@{}}$0.263\pm0.089$ \\
    $0.169\pm0.077$\end{tabular} \\
    \midrule
    0.75 &
    \begin{tabular}{@{}c@{}}$0.231\pm0.080$\\
    $0.157\pm0.062$\end{tabular} & 
    \begin{tabular}{@{}c@{}}$0.261\pm0.091$ \\
    $0.184\pm0.063$\end{tabular} &
    \begin{tabular}{@{}c@{}}$0.237\pm0.097$ \\
    $0.219\pm0.075$\end{tabular} \\
    \midrule
    1 &
    \begin{tabular}{@{}c@{}}$0.191\pm0.086$\\
    $0.231\pm0.055$\end{tabular} & 
    \begin{tabular}{@{}c@{}}$0.202\pm0.094$ \\
    $0.164\pm0.059$\end{tabular} &
    \begin{tabular}{@{}c@{}}$0.228\pm0.116$ \\
    $0.124\pm0.049$\end{tabular} \\
  \bottomrule
  \end{tabular}
\end{table}

\section{Hyperparameter search}\label{appendix_hypopt}

All models were implemented on the PyTorch framework and trained with NVIDIA V100 GPUs. 

\textbf{UPSL}~~
We performed hyperparameter tuning by varying batch size ($2$, $4$, $8$, $16$, $32$, $64$) and learning rate ($10^{-2}$, $10^{-3}$, $10^{-4}$, $10^{-5}$) options, and we selected batch size $4$ and learning rate $10^{-3}$ according to the validation performance. 

\textbf{MPSL}~~
We performed the same hyperparameter search as UPSL, and selected batch size $32$ and learning rate $10^{-3}$ for MPSL. 

\textbf{PXL}~~
Apart from batch size and learning rate, we performed hyperparameter search over identity projection, linear projection and projection with $1$, $2$ or $3$ hidden layers with dimensionality identical to the representation to choose the projection head $h$. 
We evaluated five trade-off parameters ($0$, $0.25$, $0.5$, $0.75$, $1$) to balance the supervised loss and the contrastive loss. 
We also compared the Adam optimizer \citep{kingma2014adam} and the RAdam optimizer \citep{liu2019variance}. 

After hyperparameter search, we used an identity projection head, batch size $4$, learning rate $10^{-4}$, trade-off parameter $\lambda$ $0.75$ and the Adam optimizer in the PXL model.

\section{Minibatch centered kernel alignment}\label{appendix_minicka}

To understand neural network representations, recent studies have proposed various methods including canonical correlation analysis (CCA) \citep{hardoon2004canonical}, singular vector canonical correlation analysis (SVCCA) \citep{raghu2017svcca}, projection-weighted CCA (PWCCA) \citep{morcos2018insights} and centered kernel alignment (CKA) \citep{kornblith2019similarity}. Among all approaches, CKA can reliably measure similarities of representations whose dimensions are higher than the number of samples, and consistently identify correspondences between layers across different neural network architectures and initializations \citep{kornblith2019similarity}. We utilized minibatch CKA \citep{nguyen2020wide} to evaluate representational similarity because of its computational efficiency for high-dimensional neuroimaging data. 


The mechanism of minibatch CKA is described as follows. Let $\mathbf{X} \in \mathbb{R}^{m \times u_1}$ and $\mathbf{Y} \in \mathbb{R}^{m \times u_2}$ denote representations of two layers, where $m$ is the number of samples, and $u_1$ and $u_2$ are the number of neuron units in $\mathbf{X}$ and $\mathbf{Y}$, respectively. Here, $m$ is $200$ subjects in the test set. We flattened channels $c$ and three spatial dimensions (width $w$, height $h$, depth $d$) of a convolutional layer into $u$ neurons to compare representations of different layers, i.e. $u = c \times h \times w \times d$ \citep{raghu2017svcca}.
We then randomly split $m$ subjects into $k$ minibatches and each minibatch contains $n$ subjects. Let $\mathbf{X}_i \in \mathbb{R}^{n \times u_1}$ and $\mathbf{Y}_i \in \mathbb{R}^{n \times u_2}$ denote representations of two layers in the $i$th batch. We then compute the similarity matrices $\mathbf{K}=\mathbf{X}_i \mathbf{X}_i^T$ and $\mathbf{L}=\mathbf{Y}_i \mathbf{Y}_i^T$ and estimate the similarity of the similarity matrices using Hilbert-Schmidt Independence Criterion (HSIC) \citep{gretton2005measuring}.

Minibatch CKA is computed by averaging the linear CKA across $k$ minibatches:
\begingroup\makeatletter\def\f@size{7}\check@mathfonts
\begin{equation}
    \mathrm{CKA} =\frac{ \frac{1}{k} 
\sum_{i=1}^k \mathrm{H}(\mathbf{K}, \mathbf{L}) }{ \sqrt{\frac{1}{k}
\sum_{i=1}^k \mathrm{H}(\mathbf{K}, \mathbf{K})} \sqrt{\frac{1}{k} 
\sum_{i=1}^k \mathrm{H}(\mathbf{L}, \mathbf{L})} },
\end{equation}
\endgroup

An unbiased estimator of HSIC \citep{song2012feature} is used in minibatch CKA:
\begingroup\makeatletter\def\f@size{7}\check@mathfonts
\begin{equation}
    \mathrm{H}( \mathbf{K}, \mathbf{L} )=\frac{1}{n(n-3)} (\mathrm{tr}(\mathbf{\tilde{K}} \mathbf{\tilde{L}}) + \frac{ \mathbf{1}^T \mathbf{\tilde{K}} \mathbf{1} \mathbf{1}^T \mathbf{\tilde{L}} \mathbf{1}}{(n-1)(n-2)} - \frac{2}{n-2} \mathbf{1}^T \mathbf{\tilde{K}} \mathbf{\tilde{L}} \mathbf{1} ),
\end{equation}
\endgroup
where $\mathbf{\tilde{K}}$ and $\mathbf{\tilde{L}}$ are obtained by setting the diagonal entries of $\mathbf{K}$ and $\mathbf{L}$ to zero. 

We include $8$ subjects in each minibatch in our study. Note that the CKA values are independent of the selection of batch sizes because of the unbiased estimator of HSIC.
Detailed proof of the feasibility of using minibatch CKA to approximate CKA can be found in \citep{nguyen2020wide}.

\section{Experiments using the DCGAN encoder}
\label{upsl_dcgan}
\begin{figure}[htp]
      \centering
      \includegraphics[width=2.5in]{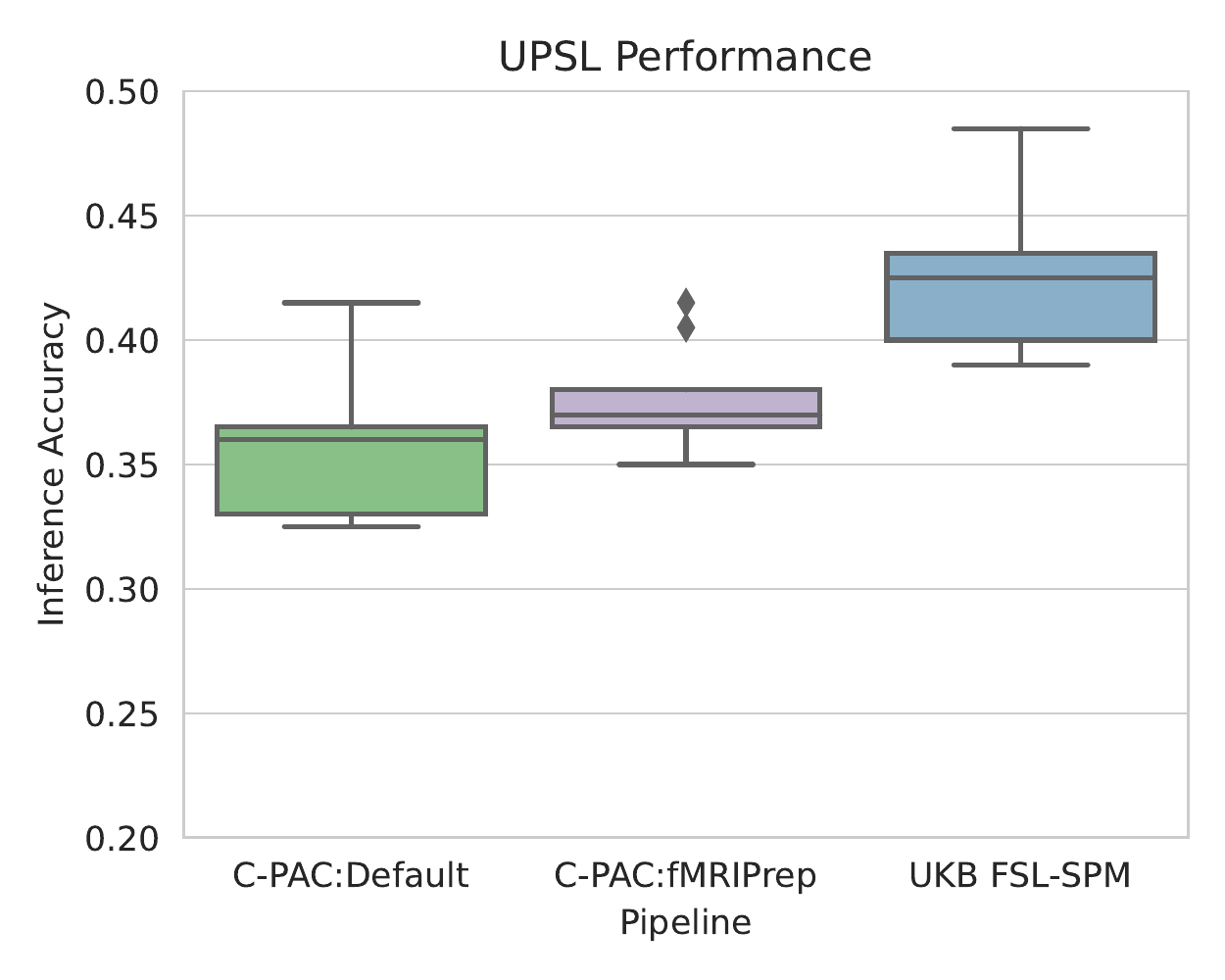}
      \caption{\textbf{UPSL within-sample inference performance using the DCGAN encoder.}}
      \label{result_upsl_dcgan}
  \end{figure}

To further verify the pipeline effect in a different encoder, we replicated the UPSL experiment using an effective unsupervised representation learning encoder -- deep convolutional generative adversarial network (DCGAN) \citep{radford2015unsupervised}. We observed a similar pipeline-related effect – the average within-sample inference accuracies across 9 folds are $35.67\%$, $37.61\%$, $42.61\%$ for C-PAC: Default, C-PAC: fMRIPrep and UKB FSL-SPM, respectively (Figure \ref{result_upsl_dcgan}). The result on the DCGAN encoder demonstrates that the pipeline-related variability exists across different encoders. 

\section{CKA results}
\label{appendix_cka}

\begin{figure}[htp]
      \centering
      \includegraphics[width=3in]{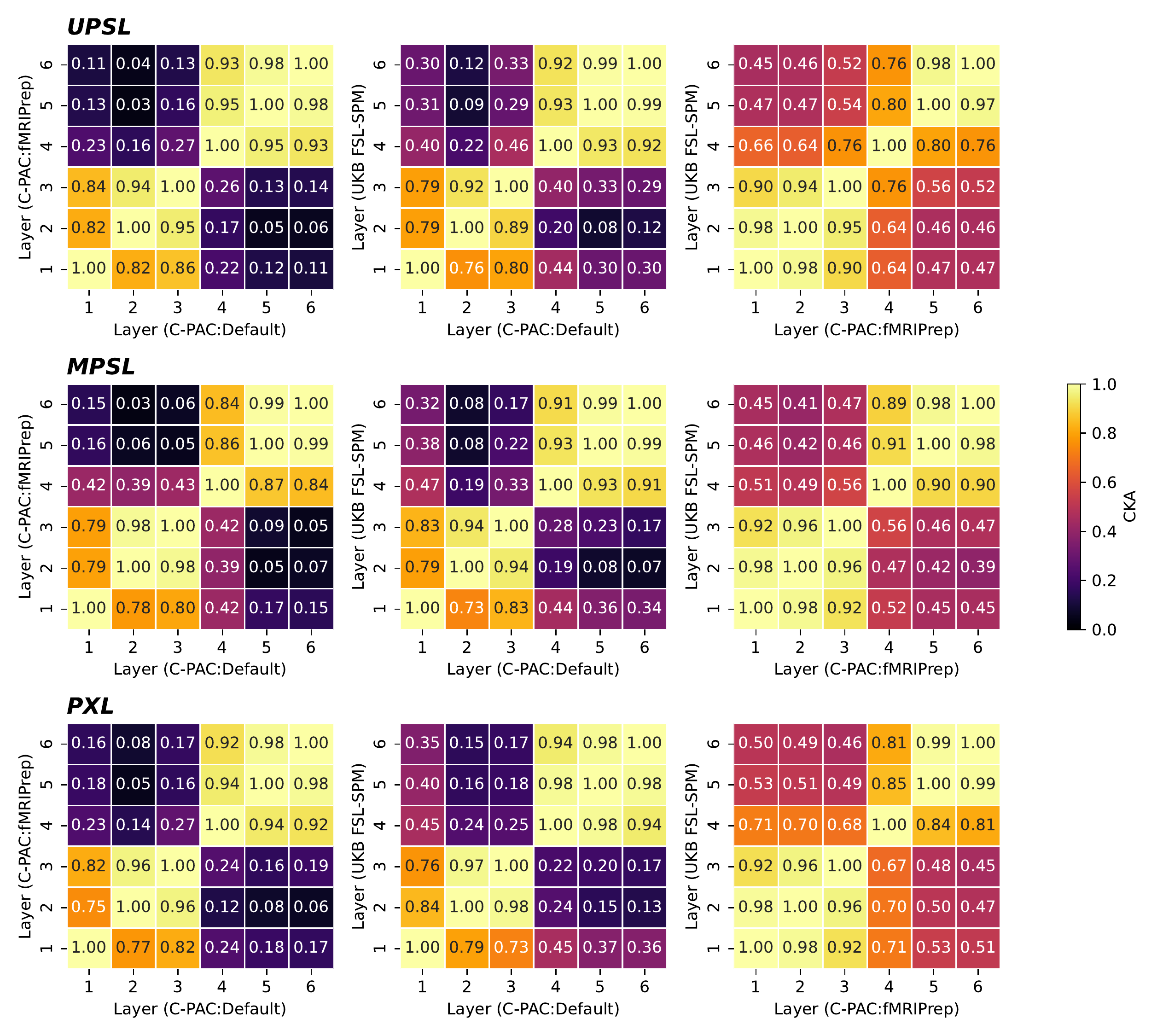}
      \caption{\textbf{Cross-layer within-pipeline CKA.}}
      \label{within_pipeline_cka}
  \end{figure}

As shown in Figure \ref{within_pipeline_cka}, the within-pipeline CKA across six layers illustrates how the neural network learns for each model. We observe that the first three layers are more similar to each other, and the last three layers are more similar, but the similarity between the first three and last three layers are low across all three learning paradigms. Table \ref{cka_table} presents the mean and the standard deviation of between-pipeline CKA values at the fourth, fifth and sixth layer and the average across these three layers. Among three models, PXL shows the highest average CKA values across the last three layers.

\begin{table}[tpb]
\tiny
  \caption{\textbf{Between-pipeline CKA.}}
  \label{cka_table}
  \begin{tabular}{cccc}
  \toprule
  Model & UPSL & MPSL & PXL  \\
    \midrule
    4th Layer & $0.652 \pm 0.071$ & $0.632 \pm 0.069$ & $\mathbf{0.771 \pm 0.043}$ \\
    5th Layer & $0.619 \pm 0.066$  & $\mathbf{0.705 \pm 0.049}$ & $\mathbf{0.705 \pm 0.047}$ \\
    6th Layer & $0.629 \pm 0.065$ & $\mathbf{0.730 \pm 0.054}$ & $0.679 \pm 0.046$ \\
    Average & $0.633 \pm 0.059$ & $0.689 \pm 0.044$  & $\mathbf{0.718 \pm 0.035}$ \\
  \bottomrule
  \end{tabular}
\end{table}

\section{Experiments on natural image datasets}
\label{appendix_mnist}

We have demonstrated that PXL can achieve consistent within-sample inference performance and capture similar cross-layer between-pipeline representations on datasets paired by subjects. To further validate the feasibility of PXL on samples paired by labels, we perform UPSL and PXL experiments on two natural image datasets, two-view MNIST and MNIST-SVHN, in which images are paired by labels. We then measure similarity  across layers using minibatch CKA.

\subsection{Datasets}
Two natural image datasets, two-view MNIST and MNIST-SVHN, are used to validate the feasibility of PXL in a broader context of multi-view and multi-domain learning. 

\textbf{Two-view MNIST}~~The two-view MNIST dataset contains two corrupted views of digits from the MNIST dataset \citep{lecun1998gradient}. 
The image preprocessing is as follows.
First of all, the intensities of each image are rescaled to a unit interval and then the size of each image is rescaled to $32 \times 32$ to fit the DCGAN architecuture.
In the first view, the image is randomly rotated at an angle uniformly sampled from $[-\frac{\pi}{4}, \frac{\pi}{4}]$.
In the second view, we add a noise sampled from a uniform distribution $[0, 1]$ to the image and additionally rescale the image intensities to a unit interval.
The cross-validation dataset is generated using stratified $5$-fold split from the original training MNIST set, including $48,000$ and $12,000$ images in the training and validation set, respectively.
The original $10,000$ MNIST test set is used as the hold-out set.

\textbf{MNIST-SVHN}~~The MNIST-SVHN dataset includes samples with two views --- grayscale MNIST digits as the first view and RGB street view house numbers sampled from the SVHN dataset \citep{netzer2011reading} as the second view. 
The dataset generation process is almost identical to the description at \url{https://github.com/iffsid/mmvae} except for two differences.
Firstly, the MNIST image is rescaled to a size of $32 \times 32$ to fit the DCGAN architecuture.
Secondly, the original training set is used to generate cross-validation folds using stratified $5$-fold split.
To generate pairs in each split, each instance of a digit class in one dataset is randomly paired with 20 instances of the same digit class from the other dataset.

\subsection{Model Architecture}
DCGAN \citep{radford2015unsupervised} is used as the encoder in the natural image experiments. The DCGAN encoder includes $4$ convolutional layers. The $4$ convolutional layers have $32$, $64$, $128$, $64$ output units, and $16^2$, $8^2$, $4^2$, $1^2$ output dimensions, respectively.
The model is trained for $200$ epochs using the RAdam optimizer with a batch size of $64$ and a learning rate of $0.0004$.

\subsection{Results}
We first train a UPSL model on each view of each dataset independently. We then train a PXL model on each two-view dataset, respectively. We repeat the experiments using $5$-fold cross-validation. 

As presented in Table \ref{appendix_table_infer}, both UPSL and PXL show comparable within-sample inference performance in general. UPSL performance is slightly better than PXL on two-view MNIST while PXL is better on MNIST-SVHN. Interestingly, we are able to achieve nearly perfect inference performance on MNIST regardless of learning paradigms, but the performance on SVHN is not ideal. One possibility can be that the SVHN image is an RGB image with three channels while the MNIST image is in grayscale with only one channel. Also note that the inference performance on SVHN improves from $0.860$ in UPSL to $0.868$ in PXL, suggesting PXL has the potential to achieve optimal result in a more challenging task.

\begin{figure}[thpb]
      \centering
      \includegraphics[width=3in]{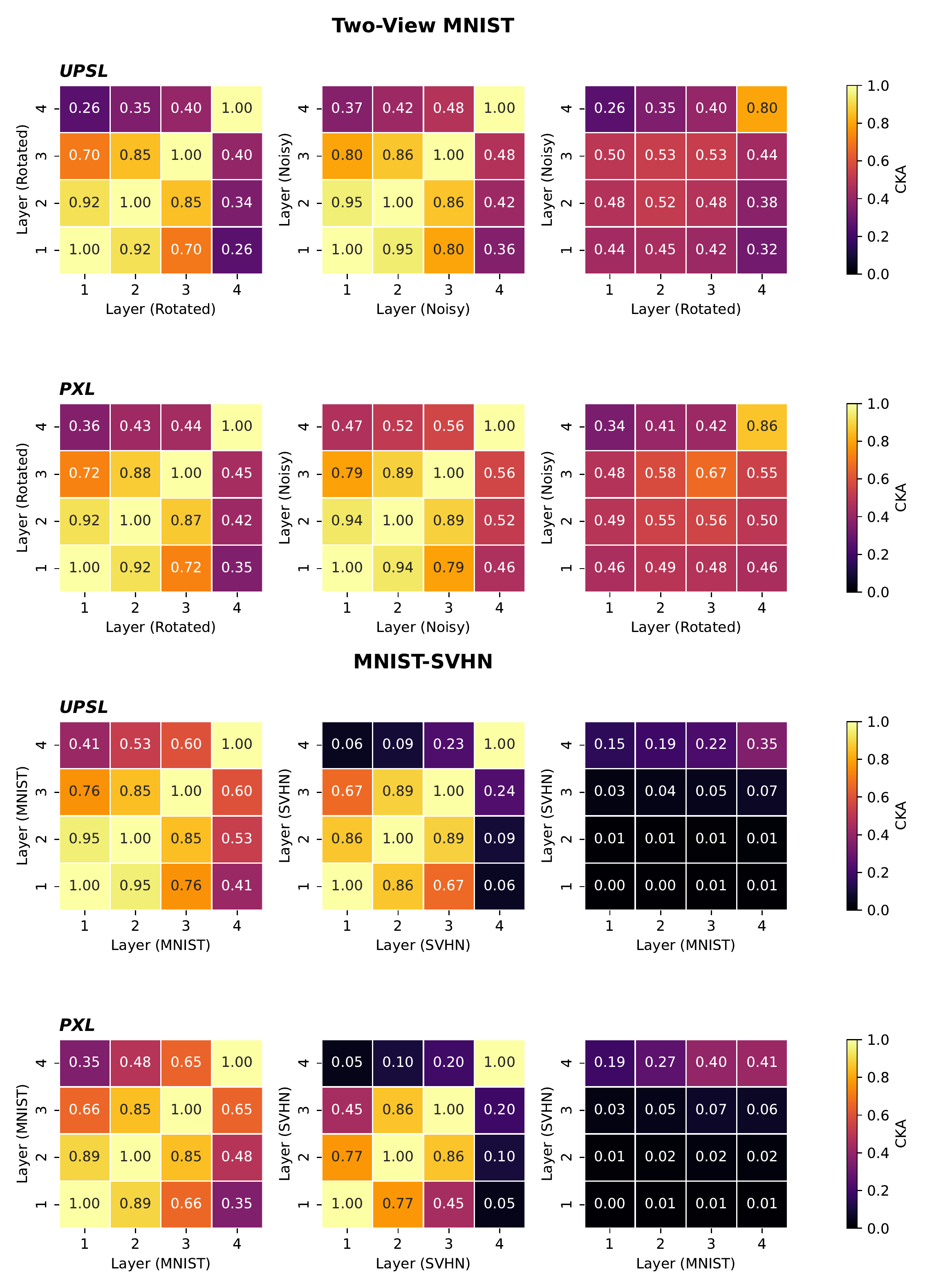}
      \caption{\textbf{CKA on natural images.} The rows correspond to UPSL and PXL CKA on two-view MNIST dataset, and UPSL and PXL CKA on MNIST-SVHN dataset correspondingly. The left two columns represent within-view CKA while the right column shows between-view CKA.}
      \label{appendix_cka_mnist}
   \end{figure}

From CKA result in Figure \ref{appendix_cka_mnist}, we observe that PXL improves between-view representational similarity on both datasets, empirically demonstrating the feasibility of PXL to capture invariant representations between different views in latent space. Thus we have shown that PXL can improve the representational similarity compared to UPSL on datasets paired by labels.


\begin{table}[tpb]
\tiny
  \caption{\textbf{Within-sample inference performance.} UPSL and PXL show comparable within-sample inference performance. Interestingly, PXL is slightly better than UPSL on SVHN test set.}
  \label{appendix_table_infer}
  \begin{tabular}{cccc}
  \toprule
    Training Set & Model & Rotated & Noisy \\
    \midrule
    Two-view MNIST & UPSL & $0.991 \pm 0.000$ & $0.989 \pm 0.000$ \\
    Two-view MNIST & PXL & $0.987 \pm 0.011$ & $0.984 \pm 0.009$ \\
    \bottomrule
    \toprule
    Training Set & Model & MNIST & SVHN \\
    \midrule
    MNIST-SVHN & UPSL & $0.989 \pm 0.001$ & $0.860 \pm 0.003$ \\
    MNIST-SVHN & PXL & $0.991 \pm 0.001$ & $0.868 \pm 0.001$ \\
    \bottomrule
  \end{tabular}
\end{table}

\end{document}